%% file: main.tex
\documentclass[10pt]{article} %
\usepackage[preprint]{tmlr}

\input{math_commands.tex}

\usepackage{hyperref}       %
\usepackage{url}            %
\usepackage{booktabs}       %
\usepackage{amsfonts}       %
\usepackage{nicefrac}       %
\usepackage{xcolor}         %
\usepackage{adjustbox}
\usepackage{amsmath}
\usepackage{amssymb}
\usepackage{bm}
\usepackage{colortbl}
\usepackage{comment}
\usepackage{dsfont}
\usepackage{enumitem}
\usepackage{graphicx} 
\usepackage{mathtools}
\usepackage{multicol}
\usepackage{multirow}
\usepackage{lscape}
\usepackage{pifont}
\usepackage{scalerel}
\usepackage{soul}
\usepackage{subcaption}
\usepackage{tabularx}
\usepackage[most]{tcolorbox}
\usepackage{xspace}
\usepackage{wrapfig}
\usepackage{algorithm}
\usepackage{algpseudocode}
\usepackage{tablefootnote}

\definecolor{mygreen}{HTML}{009B55}
\hypersetup{colorlinks=true,urlcolor=mygreen,linkcolor=mygreen,citecolor=mygreen}

\newcommand{\diffvec}{diff vector\xspace}
\newcommand{\emphdiffvec}{\emph{diff vector}\xspace}

\newcommand{\smallurl}[1]{ \begin{scriptsize}\url{#1}\end{scriptsize}}
\newcommand{\smallsup}[1]{\scaleto{\text{#1}}{4pt}}
\newcommand{\plcomment}[1]{\textcolor{black}{#1}}%

\newcommand{\rbcomment}[1]{\textcolor{black}{#1}}

\newcommand{\tvcomment}[1]{\textcolor{black}{#1}}

\title{Efficient Model Development through Fine-tuning Transfer}

\author{}
\author{\name Pin-Jie Lin$^{1}$ \email pinjie@vt.edu
  \AND
  \name Rishab Balasubramanian$^{1}$ \email rishbb@vt.edu
  \AND
  \name Fengyuan Liu$^{2}$ \email fy.liu@mail.utoronto.ca
  \AND
  \name Nikhil Kandpal$^{2}$ \email nkandpa2@cs.toronto.edu
  \AND
  \name Tu Vu$^{1}$ \email tuvu@vt.edu
  \AND \addr $^{1}$Virginia Tech ~~ $^{2}$University of Toronto \& Vector Institute
}

\begin{document}
\maketitle
\input{sections/abstract}
\input{sections/introduction}

\input{sections/recycling_finetuning}
\input{sections/multilingual_model_development}

\input{sections/controlled_experiments}

\input{sections/recycling_then_finetuning}

\input{sections/iterative_recycling_then_finetuning}

\input{sections/related_work}

\input{sections/conclusion}

\input{sections/acknowledgements}
\newpage
\bibliography{custom}
\bibliographystyle{tmlr}
\newpage
\input{sections/appendix}

\end{document}

%% file: math_commands.tex
\usepackage{amsmath,amsfonts,bm}

\def\eqref#1{equation~\ref{#1}}

\def\1{\bm{1}}

\DeclareMathAlphabet{\mathsfit}{\encodingdefault}{\sfdefault}{m}{sl}
\SetMathAlphabet{\mathsfit}{bold}{\encodingdefault}{\sfdefault}{bx}{n}

%% file: sections/abstract.tex
\begin{abstract}
\plcomment{Modern LLMs struggle with efficient updates, as each new pretrained model version requires repeating expensive alignment processes. This challenge also applies to domain- or language-specific models, where fine-tuning on specialized data must be redone for every new base model release. In this paper, we explore the transfer of fine-tuning updates between model versions. Specifically, we derive the \emph{diff vector} (representing the weight changes from fine-tuning) from one \emph{source} model version and apply it to the base model of a different \emph{target} version. Through empirical evaluations on various open-weight model versions, we show that transferring diff vectors can significantly improve the performance of the target base model. For example, transferring the fine-tuning updates from Llama 3.0 8B improves Llama 3.1 8B by 46.9\% on IFEval and 15.7\% on LiveCodeBench without additional training, even surpassing Llama 3.1 8B Instruct. Furthermore, we demonstrate performance gains on multilingual tasks, with 4.7\% and 15.5\% improvements on Global MMLU for Malagasy and Turkish, respectively. We observe that these merged models provide stronger initializations for further fine-tuning. Lastly, our controlled experiments suggest that fine-tuning transfer is most effective when source and target models lie in a linearly connected region of parameter space, and we provide a theoretical analysis of our method. Taken together, fine-tuning transfer offers a cost-efficient and practical strategy for continuous LLM development. Our code is available at \href{https://github.com/pjlintw/finetuning-transfer}{github.com/pjlintw/finetuning-transfer}.}
\end{abstract}

%% file: sections/introduction.tex
\section{Introduction}
\input{figures/recycling_pipeline}
\plcomment{Today's large language models (LLMs) are developed in two stages: (1) \emph{pretraining} on massive corpora with self-supervised learning, and (2) \emph{post-training} with alignment steps~\citep{ouyang2022training, bai2022training}. While this pipeline creates powerful LLMs, it presents a major bottleneck for continuous development: every new version of a pretrained model requires repeating expensive post-training. This challenge is particularly acute in domain- or language-specific applications, where the cost of redoing fine-tuning for each base model update is prohibitive~\citep{qin-etal-2023-recyclable, bandarkar2024layer}.}

\plcomment{In this paper, we explore a method to reduce post-training costs by transferring fine-tuning updates between different model versions. Specifically, we propose incorporating the \emph{weight updates} from a \emph{source} model version $s$ to improve a \emph{target} model version $t$.} Our approach (see Figure~\ref{figure:recycling_pipeline}) first computes the \emphdiffvec $\Delta_s = m'_{s} - m_s$ from version $s$, which represents the difference between the fine-tuned model $m'_{s}$ (e.g., instruction-tuned) and its base model $m_s$ (pretrained). 
Intuitively, $\Delta_s$ encodes the task-specific updates to the model parameters during fine-tuning, and can be used to transfer knowledge from the source version $s$ to the target version $t$. \plcomment{Contrary to prior work~\citep{ilharco2023editing,huang2023chat}, which focuses on improving the capabilities of a single model on a specific target task, we focus on a general-purpose method to transfer updates between different model versions for a variety of downstream tasks.} We hypothesize that models fine-tuned using the same or similar training data and procedures exhibit linear relationships across versions: $m'_{s} - m_{s} \approx m'_{t} - m_{t}$. This suggests that we can approximate the fine-tuned version $m'_{t}$ of the target base model $m_{t}$ without training: $m'_{t} \approx m_{t} + \Delta_{s}$. \plcomment{The intuition is supported by linear mode connectivity theory~\citep{mirzadeh2020linear,pmlr-v119-frankle20a}, which shows that two independently trained networks can be connected by a low-loss path (see Appendix~\ref{appendix:theoretical_justification}).}

We begin by evaluating the feasibility of our approach through the transfer of \emph{diff vectors} across different versions of open-weight models (Section~\ref{section:recycling_finetuning}). Recycling the fine-tuning updates from Llama 3.0 yields a 46.9\% absolute accuracy improvement on IFEval over Llama 3.1 8B, while also surpassing the performance of Llama 3.1 8B Instruct without additional training.

Motivated by these results, we conduct a case study on the development of multilingual models (Section~\ref{section:multilingual_model_development}). 
We observe that diff vectors transfer facilitates a better understanding of the target language. Specifically, transferring weights from a fine-tuned version of Llama 3.0 Instruct to Llama 3.1 Instruct yields absolute accuracy improvements of 4.7\% for Malagasy and 15.5\% for Turkish on the Global MMLU benchmark~\citep{singh2024global}, without additional training.

To shed light on when fine-tuning transfer is most effective, we perform controlled experiments using OLMo 2's~\citep{olmo20242} intermediate pretrained checkpoints as different model versions (Section \ref{section:controlled_experiments}). Our results suggest that fine-tuning transfer is most effective when the source and target models lie within a linearly connected region of the parameter space, \plcomment{consistent with} linear mode connectivity~\citep{mirzadeh2020linear,ainsworth2023git,pmlr-v162-wortsman22a,Wortsman_2022_CVPR,pmlr-v119-frankle20a}.

Furthermore, we investigate whether the merged model $m_{t} + \Delta_{s}$ can serve as a computationally efficient and effective starting point for fine-tuning (Section \ref{section:fine_tuning_transfer_as_starting_point}). Our experiments demonstrate that initializing fine-tuning with this merged model can accelerate convergence and improve accuracy compared to training on top of $m_t$. \plcomment{We find that even when the selected diff vector is suboptimal, fine-tuning the merged model consistently improves performance compared to direct fine-tuning, without harming generalization to unseen tasks. This suggests that fine-tuning transfer can serve as a robust and effective intermediate step when training is feasible.} 

Lastly, we explore a continuous model development scenario in which new model versions are regularly released (Section~\ref{section:iterative_recycling_then_finetuning}). We propose an iterative recycling–then–fine-tuning approach that incrementally accumulates fine-tuning updates from previous versions.

In summary, our key contributions are as follows.
\begin{itemize}
    \item Introducing an approach for transferring fine-tuning updates between model versions via diff vector transfer.
    \item Demonstrating that this approach can reduce training costs while maintaining competitive performance.
    \item Validating the approach in a multilingual model development setting, showing improved language-specific performance without retraining.
    \item Establishing conditions for effective fine-tuning transfer, particularly when models exhibit linear mode connectivity. 
    \item Proposing a recycling-then-finetuning strategy to improve both efficiency and performance in a continuous model development setting.
\end{itemize}

\paragraph{Terminology:} In this paper, we use the terms \emph{``transferring fine-tuning''}, \emph{``transferring fine-tuning updates''} interchangeably to refer to the process of reusing fine-tuning updates (i.e., weight changes) from a source model version and incorporating them into a target version.

%% file: figures/recycling_pipeline.tex
\begin{figure}[t]
\centering
\includegraphics[width=0.65\textwidth]{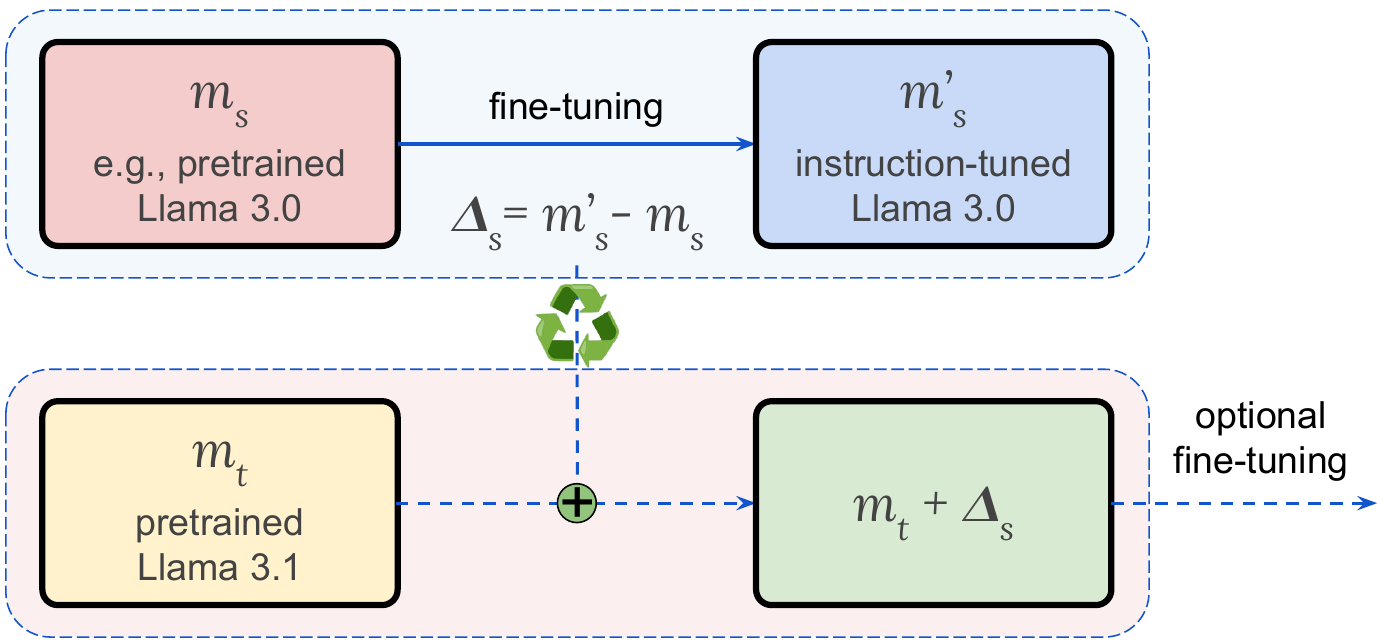}
\caption{\plcomment{To transfer fine-tuning (e.g., instruction tuning) from a \emph{source} model version $s$ (e.g., Llama 3.0) to a \emph{target} version $t$ (Llama 3.1), we first compute the \diffvec $\Delta_s = m'_{s} - m_s$ from version $s$, where  $m'_{s}$ is the fine-tuned model (instruction-tuned Llama 3.0) and $m_s$ is the base model (pretrained Llama 3.0). Then, we add $\Delta_s$ to the target base model (pretrained Llama 3.1) to approximate the fine-tuned model in version $t$ (instruction-tuned Llama 3.1).}}
\label{figure:recycling_pipeline}
\end{figure}

%% file: sections/recycling_finetuning.tex
\section{Transferring fine-tuning updates across model versions} 
\label{section:recycling_finetuning}
In this section, we explore transferring the weight changes from a source model version $s$ to a target model version $t$, denoted $\mathcal{T}_{s \rightarrow t}$, without additional training. Specifically, we directly merge (add) the \diffvec $\Delta_s = m'_{s} - m_s$ from version $s$, which captures the parameter adaptations from the base model $m_s$ to its fine-tuned counterpart $m'_{s}$, onto the new base model $m_t$ in version $t$, without any gradient-based training. Our results (Table~\ref{table:recyling_finetuning}) show that fine-tuning updates can be effectively transferred across model versions, as $m_t + \Delta_s$ often performs comparably to its fine-tuned counterpart $m'_{t}$.

\subsection{Experimental setup}
\label{subsection_transfer_finetuning_experimental_setup}
We conduct experiments with various open-weight models, including Llama~\citep{dubey2024llama}, OLMo~\citep{olmo20242}, and T\"ulu~\citep{lambert2024t}. 
\plcomment{Throughout this work, we ensure that our source and target models are of the same architecture. We provide additional cross-architecture transfer results in Appendix \ref{appendix:recycling_finetuning_across_architectures} and leave further research on cross-architecture recycling as future work.}
Our study explores both transfer directions: from an older model version to a newer one (\emph{recycling}) and from a newer version to an older one (\emph{backporting}). 

\input{tables/recycling_finetuning}

\emph{Recycling} can save training time and computational resources, while incorporating post-training capabilities into the newer pretrained model.
Conversely, \emph{backporting} is beneficial when the older base model is better optimized for a specific use case (e.g., a particular language), allowing the user to take advantage of the new fine-tuning improvements while maintaining optimization and compatibility.\footnotemark
\footnotetext{In software development, \emph{backporting} refers to the process of adapting features or updates from a newer version of a software system or component for use in an older version.} We emphasize that our goal is not to achieve state-of-the-art results, but instead to assess the feasibility of transferring fine-tuning updates between model versions.

We evaluate the merged model $m_t + \Delta_s$ on a diverse set of benchmarks, including general knowledge with MMLU~\citep{hendrycks2021ameasuring}, math with GSM8K~\citep{cobbe2021training} and MATH~\citep{hendrycks2021bmeasuring}, reasoning with $\text{ARC}_\text{C}$~\citep{clark2018think} and GPQA~\citep{rein2024gpqa}, instruction-following with IFEval~\citep{zhou2023instruction}, \plcomment{code generation with HumanEval+ (HE+ in Table \ref{table:recyling_finetuning}}) and MBPP+~\citep{liu2023your}, LiveCodeBench~\citep{jain2024livecodebench}, and BigCodeBench~\citep{zhuo2024bigcodebench} (LCB and BCB in Table \ref{table:recyling_finetuning} respectively). 
We compare its performance to that of directly fine-tuned $m_t$ (i.e., $m'_{t}$).\footnote{\plcomment{For evaluating HumanEval+ and MBPP+ we use \texttt{EvalPlus}~\citep{liu2023your}, and the official evaluation libraries for LiveCodeBench and BigCodeBench. All other tasks are evaluated using the \texttt{lm-evaluation-harness} library~\citep{eval-harness}. }} See Appendix~\ref{appendix_evaluation_details} for evaluation details.

\subsection{Results and discussion}
\paragraph{Transferring fine-tuning substantially boosts the target base model's performance:} Table~\ref{table:recyling_finetuning} shows our results when transferring fine-tuning (i.e., instruction tuning) updates between Llama 3.0 and Llama 3.1. \plcomment{First, we note that Llama 3.0 Instruct consistently performs better than Llama 3.1 (and vice versa). This highlights that most capabilities of the instruction-tuned model arise post-training. Here, we attempt to transfer such capabilities between model versions, and thus bypass the alignment stage.} Strikingly, adding the diff vector $\Delta_s$ from a different model version can effectively transform a non-instruction-tuned model (e.g., Llama 3.0 or Llama 3.1) into one that follows instructions well (Llama 3.0 + $\Delta_{3.1}$ or Llama 3.1 + $\Delta_{3.0}$). For example, our approach yields 42.1\% and 46.9\% absolute accuracy improvements on the instruction-following benchmark IFEval over the base versions of Llama 3.0 and Llama 3.1, respectively. Large gains are also observed across the board on math, \rbcomment{code,} and reasoning benchmarks, \rbcomment{with an average improvement of 16.5\% for Llama 3.0 and 14.4\% for Llama 3.1.}
These results suggest that advanced knowledge and instruction-following abilities can be efficiently transferred between model versions without further training. In general, Llama 3.0 benefits more from the backported \diffvec $\Delta_{3.1}$ from version 3.1 than Llama 3.1 does from recycling version 3.0's \diffvec $\Delta_{3.0}$.

\paragraph{Transferring fine-tuning can achieve performance comparable to the fine-tuned model:} Our results demonstrate that the merged model $m_t + \Delta_s$ can perform on par with its fine-tuned counterpart $m'_{t}$ across various tasks. This is particularly true for Llama 3.0 + $\Delta_{3.1}$, which matches or surpasses Llama 3.0 Instruct on \rbcomment{eight out of ten} tasks we evaluated. Interestingly, Llama 3.1 + $\Delta_{3.0}$ outperforms LLama 3.1 Instruct on \rbcomment{four out of the ten} benchmarks. This is a testament to the \diffvec's ability to encode advanced reasoning and instruction-following capabilities. Overall, our results suggest that fine-tuning transfer provides an effective and extremely low-cost method to improve model performance when training is prohibitively expensive.

\paragraph{Transferring fine-tuning can induce step-by-step reasoning:}
\tvcomment{Interestingly, we observed that transferring fine-tuning updates consistently shifts the target base model's answers from direct responses to step-by-step reasoning (Appendix~\ref{appendix:generation_results_gsm8k_math}). This emergent reasoning behavior appears after adding the diff vector and aligns with the accuracy improvements on GSM8K and MATH benchmarks.}

%% file: tables/recycling_finetuning.tex
\begin{table*}[t]
\centering
\begin{adjustbox}{max width=\textwidth}
\begin{tabular}{lccccccccccc}
\toprule
$\mathbf{Model}$ & $\mathbf{GSM8K}$ & $\mathbf{MATH}$ & $\mathbf{ARC_C}$ & $\mathbf{GPQA}$ & $\mathbf{MMLU}$ & $ \mathbf{IFEval}$ & \plcomment{$\mathbf{HE+}$} &  \plcomment{$\mathbf{MBPP+}$} & \plcomment{$\mathbf{LCB}$} & \plcomment{$\mathbf{BCB}$} & \plcomment{\textbf{Avg.}}\\
\midrule
\rowcolor{gray!10} Llama 3.0 8B Instruct & 81.1 & 28.8  & 82.4 & \textbf{31.5} & 64.9 & \textbf{76.6} & 56.7 & \textbf{55.6} & 14.0 & 6.8 & 49.8 \\
Llama 3.0 8B & 55.6 & 17.3  & 79.7 & 22.3 & 66.7 & 34.5 & 31.1 & 51.3 & 0.0 & 6.1  & 36.5 \\
\; + $\Delta_{3.1}$ & \textbf{82.8} & \textbf{44.7} & \textbf{83.0} & 25.9 & \textbf{70.0} & \textbf{76.6} & \textbf{62.8} & {55.3} & \textbf{15.8} & \textbf{12.8} & \textbf{53.0} \\
\midrule
\rowcolor{gray!10} Llama 3.1 8B Instruct  & \textbf{86.5} & \textbf{50.3} & \textbf{83.8} & 31.3 & \textbf{72.9} & 80.5 & \textbf{61.0} & 54.8 & 16.0 & \textbf{14.9}  & \textbf{55.2} \\
Llama 3.1 8B & 56.6 & 19.3 & 79.2 & 21.9 & 66.8 & 36.4 & 29.9 & 51.9  & 0.4 & 5.4 & 36.8 \\
\; + $\Delta_{3.0}$ & 79.8 & 29.9  & 82.9 & \textbf{32.6} & 65.1 & \textbf{83.3} & 55.5 & \textbf{56.6} & \textbf{16.1} & 10.1 & 51.2 \\
\bottomrule
\end{tabular}
\end{adjustbox}
\caption{Fine-tuning transfer significantly improves the performance of the target base model across various tasks, achieving results comparable to its fine-tuned counterpart in many cases. Here, $\Delta_{3.0}$ and $\Delta_{3.1}$ represent the diff vectors between Llama Instruct and Llama for versions 3.0 and 3.1, respectively. \plcomment{HE+ stands for HumanEval+, LCB for LiveCodeBench, and BCB for BigCodeBench.} \hl{Notably, adding the diff vector $\Delta_s$ from a different model version can effectively transform a non-instruction-tuned model (e.g., Llama 3.0 or Llama 3.1) into one that follows instructions well (Llama 3.0 + $\Delta_{3.1}$ or Llama 3.1 + $\Delta_{3.0}$) without further training.} Additional results for OLMo and T\"ulu can be found in Appendix~\ref{appendix:additional_results_tulu_olmo}, \hl{where we additionally find that advanced LLM capabilities, attained through alignment tuning stages such as Supervised Fine-Tuning (SFT), Direct Preference Optimization (DPO), or Group Relative Policy Optimization (GRPO), can be successfully transferred across different model versions.}}
\label{table:recyling_finetuning}
\end{table*}

%% file: sections/multilingual_model_development.tex
\section{Efficient multilingual model development}
\label{section:multilingual_model_development}
Motivated by our results in Section~\ref{section:recycling_finetuning}, we now turn toward applying our fine-tuning transfer approach in a multilingual model development setting. We focus exclusively on a \emph{recycling} scenario, where our aim is to transfer the language-specific instruction tuning updates from an older model version to a newer one.

For language-specific instruction tuning, we fine-tune an instruction-tuned model rather than a pretrained one. This approach aligns with the common practice of using an instruction-tuned English or multilingual model as the foundation when developing language-specific models. A key challenge in this setting is that state-of-the-art LLMs often include multilingual data in pretraining and instruction tuning, which makes it unclear whether language-specific fine-tuning is still necessary. \emph{How effective is our recycling approach when applied to a multilingual instruction-tuned model?} Our results show that recycling fine-tuning remains effective in this scenario, as long as the target base model is outperformed by the fine-tuned model of the source version.

\subsection{Experimental setup}
\label{section:training_details}
We fine-tune Llama 3.0 Instruct ($m_s$) separately on language-specific instruction tuning data for three languages: Malagasy, Sinhala, and Turkish. We use the Aya dataset~\citep{singh-etal-2024-aya} for Malagasy (14.6K examples) and Sinhala (14.5K examples), and the InstrucTurca dataset~\citep{instructurca} for Turkish (16.7K examples).\footnote{To simulate a low-resource setting, we sampled 6.5\% of the original InstrucTurca dataset, which contains 2.58 million examples, resulting in approximately 16.7K examples.} Each model is trained for 30K training steps with a learning rate of 5e-6 and a batch size of 8, using 4 NVIDIA A100-80G GPUs.\footnote{We use the AdamW optimizer with a linear scheduler and a warmup ratio of 0.03. We disable dropout and exclude weight decay for embeddings. The sequence length is 2048. We use \texttt{open-instruct}~\citep{lambert2024t} for training and \texttt{lm-evaluation-harness}~\citep{eval-harness} for evaluation.} 

After training on each language, we compute the diff vector $\Delta_s = m'_s - m_s$ and add it to Llama 3.1 Instruct $m_t$. We simulate a low-resource setting and do not perform any additional training with language-specific data. The merged model $m_t + \Delta_s$ is evaluated against the base model $m_t$ on the Global MMLU benchmark~\citep{singh2024global}.

\subsection{Results and discussion}
\input{tables/controlled_multilingual_model_development}
\paragraph{Transferring fine-tuning is effective for developing multilingual models:} Our results in Table~\ref {table:recyling_finetuning_multilingual} demonstrate the benefits of reusing fine-tuning updates in multilingual model development. For Malagasy and Turkish, transferring the \diffvec from Llama version 3.0 to 3.1 results in significant accuracy improvements
(4.7\% and 15.5\%, respectively) over Llama 3.1 8B Instruct. Our recycling approach performs better than the fine-tuned Llama 3.0 Instruct model for Malagasy (1.5\% accuracy improvement) and maintains similar performance for Turkish. 

On the other hand, for Sinhala, recycling fine-tuning offers no advantage, as Llama 3.1 Instruct already outperforms the previously fine-tuned Llama 3.0 Instruct. However, even in this case, recycling does not significantly reduce performance.

%% file: tables/controlled_multilingual_model_development.tex
\begin{table}[t]
\centering
\begin{adjustbox}{max width=\textwidth}
\begin{tabular}{lccc}
\toprule
$\mathbf{Model}$ & $\mathbf{Malagasy}$ & $\mathbf{Sinhala}$ & $\mathbf{Turkish}$ \\
\midrule
\rowcolor{gray!10} Llama 3.0 8B Instruct & 23.1 & 23.3 & 30.8\\
\; + FT & 30.8 & 29.0 & 43.2\\
\midrule
\rowcolor{gray!10} Llama 3.1 8B Instruct & 27.6 & \textbf{33.0} & 27.7 \\
\; + $\Delta_{3.0}$ & \textbf{32.3} & 32.3 & \textbf{43.2}\\
\bottomrule
\end{tabular}
\end{adjustbox}
\caption{\hl{Recycling fine-tuning updates improves multilingual performance on Global MMLU without retraining, yielding a 4.7\% and 15.5\% absolute improvement for Malagasy and Turkish, respectively, compared to Llama 3.1 8B Instruct.} $\Delta_{3.0}$ represents the diff vector between Llama 3.0 Instruct and its monolingual fine-tuned (FT) version.}
\label{table:recyling_finetuning_multilingual}
\end{table}

%% file: sections/controlled_experiments.tex
\section{When is fine-tuning transfer effective?}
\label{section:controlled_experiments}
Having demonstrated the effectiveness of fine-tuning transfer, we now conduct controlled experiments to better understand when this approach is most effective. At a high level, we treat different checkpoints of a pretrained model as distinct model versions. We then fine-tune these model versions on the same data and assess the impact of transferring fine-tuning updates between them. Our results reveal that fine-tuning transfer is most successful when the source and target models are close within a linearly connected region of the parameter space, \plcomment{consistent with} linear mode connectivity. \plcomment{We provide further theoretical analysis in Appendix ~\ref{appendix:theoretical_justification}.}

\subsection{Experimental setup}
\label{section:controlled_experiment_setup}
We conduct experiments with the publicly available intermediate checkpoints of OLMo 2 7B.\footnote{\smallurl{https://huggingface.co/allenai/OLMo-2-1124-7B}} The base OLMo 2 model was trained in two stages: (1) a general web-based pretraining stage (stage 1), and (2) a mid-training stage (stage 2) using high-quality web data and domain-specific data to enhance STEM-related capabilities. We select five checkpoints: $\mathcal{M}_1$ (early-stage 1, at 300K steps), $\mathcal{M}_2$ (mid-stage 1, at 600K steps), $\mathcal{M}_3$ (end-stage 1, at 929K steps), $\mathcal{M}_4$ (mid-stage 2, at 6K steps), and $\mathcal{M}_5$ (end-stage 2, at 12K steps). Each $\mathcal{M}_i$ is treated as a distinct model version. We investigate both transfer scenarios: (1) recycling 
($\mathcal{T}_{\mathcal{M}_i \rightarrow \mathcal{M}_j}, i < j$), and (2) backporting 
($\mathcal{T}_{\mathcal{M}_j \rightarrow \mathcal{M}_i}, j > i$).

Due to our limited computational resources, supervised fine-tuning with a large instruction tuning dataset would be prohibitively expensive. We therefore fine-tune all model versions using a subset of the math reasoning instruction tuning data from T\"ulu 3, which includes Tülu 3 Persona MATH, GSM, and Algebra (220K examples total), following the training procedure described in Section~\ref{section:training_details}.

We evaluate our models on GSM8K and the MATH500 subset~\citep{hendrycks2021bmeasuring} of the MATH dataset.
These datasets are selected because fine-tuning on T\"ulu 3's math reasoning data significantly improves performance on them, allowing for a clearer analysis of the impact of transferring fine-tuning updates between model versions.\footnote{For evaluation, we use the \texttt{OLMES} library~\citep{gu2024olmes}.}

\subsection{Results and discussion}
\paragraph{More powerful models are better at leveraging transferred fine-tuning:} Our results in Table~\ref{table:controlled_recyling_finetuning_gsm8k} indicate that stronger models are more effective at leveraging transferred fine-tuning updates. While transferring fine-tuning can improve performance for $\mathcal{M}_1$, $\mathcal{M}_2$, and $\mathcal{M}_3$, the merged models $\mathcal{M}_i + \Delta_{j}$ ($\Delta_j$ denotes the \diffvec from model version $\mathcal{M}_j$, $j \neq i$) still fall significantly short of their fine-tuned counterparts, denoted FT$(\mathcal{M}_i)$. On GSM8K, the accuracy gaps between the best $\mathcal{M}_i + \Delta_{j}$ and FT$(\mathcal{M}_i)$ are 26.1\%, 24.1\%, 20.6\% for $\mathcal{M}_1$, $\mathcal{M}_2$, and $\mathcal{M}_3$, respectively. In contrast, for $\mathcal{M}_4$, this gap narrows to 2.8\%. Notably, recycling fine-tuning from $\mathcal{M}_4$ to $\mathcal{M}_5$ (i.e., $\mathcal{M}_5 + \Delta_{4}$) surpasses fine-tuning directly on $\mathcal{M}_{5}$ (FT($\mathcal{M}_5$)), achieving 1.6\% accuracy improvement (77.1\% vs. 75.5\%). Similar trends are observed on MATH500. This pattern suggests an emergent ability---effective use of transferred fine-tuning only emerges when the target base model is sufficiently strong. In other words, the benefits of transferring fine-tuning only become significant beyond a certain level of capability.
\input{tables/controlled_recycling_finetuning_gsm8k}
\paragraph{Fine-tuning transfer works best when models are close in the parameter space:} Our results also suggest that fine-tuning transfer is most effective when the source and target models are closely connected in the parameter space. On both GSM8K and MATH500, models $\mathcal{M}_1$ and $\mathcal{M}_2$ benefit more from $\Delta_3$ than from $\Delta_4$ or $\Delta_5$. Similarly, $\mathcal{M}_4$ and $\mathcal{M}_5$ gain more from $\Delta_3$ than from $\Delta_1$ or $\Delta_2$. Overall, $\mathcal{M}_1$, $\mathcal{M}_2$, and $\mathcal{M}_3$ form a mutually beneficial group, as do $\mathcal{M}_4$ and $\mathcal{M}_5$. However, transferring between these two groups can degrade performance. Specifically, $\mathcal{M}_1$, $\mathcal{M}_2$, and $\mathcal{M}_3$ do not benefit from $\Delta_4$ and $\Delta_5$, while $\mathcal{M}_4$ and $\mathcal{M}_5$ typically benefit only from $\Delta_3$.\footnote{The only exception is $\mathcal{M}_4$ benefiting from $\mathcal{M}_1$ and $\mathcal{M}_2$ on MATH500.}

%% file: tables/controlled_recycling_finetuning_gsm8k.tex
\begin{table}[t]
\centering
\begin{adjustbox}{max width=\textwidth}
\begin{tabular}{lccccc}
\toprule
\textbf{} & \textbf{$\mathcal{M}_1$} & \textbf{$\mathcal{M}_2$} & \textbf{$\mathcal{M}_3$} & \textbf{$\mathcal{M}_4$} & \textbf{$\mathcal{M}_5$} \\
\midrule
\rowcolor{gray!10} & 13.2 & 19.4 & 24.4 & 64.5 & 65.5 \\
{\; + $\Delta_1$} &   & \textbf{26.6} & 32.0 & 27.5 & 19.6 \\
\rowcolor{gray!10}{\; + $\Delta_2$} & \textbf{19.0} &  & \textbf{39.8} & 25.9 & 17.3 \\
{\; + $\Delta_3$} & 14.3 & 25.0 &   & 68.6 & 70.3 \\
\rowcolor{gray!10}{\; + $\Delta_4$} & 11.8 & 18.0 & 22.6 & & \textbf{77.1}  \\
{\; + $\Delta_5$}  & 11.9  & 16.0 & 24.0 & \textbf{72.9} &  \\
\midrule
FT(\textbf{$\mathcal{M}_i$}) & 45.1 & 50.7 & 60.4 & 75.7 & 75.5 \\
\bottomrule
\end{tabular}
\end{adjustbox}
\caption{\hl{GSM8K accuracies indicating that more powerful models are better at leveraging transferred fine-tuning. Effective use of transferred fine-tuning only emerges once the target base model reaches a certain level of capability. Furthermore, fine-tuning transfer
works best when the source and target models are close within a linearly connected region of the parameter space.} Here, $\mathcal{M}_i$ represents different intermediate pretrained checkpoints of OLMo 2 7B (with smaller values of $i$ indicating earlier checkpoints), and $\Delta_i$ refers to the \diffvec resulting from the fine-tuning of version $i$. FT(\textbf{$\mathcal{M}_i$}) denotes applying fine-tuning directly to $\mathcal{M}_i$. See Table~\ref{appendix_table:controlled_recyling_finetuning_math500} in Appendix~\ref{appendix:additional_results_section_4} for MATH500 results.}
\label{table:controlled_recyling_finetuning_gsm8k}
\end{table}

%% file: sections/recycling_then_finetuning.tex
\section{Fine-tuning transfer as a starting point for further fine-tuning}
\label{section:fine_tuning_transfer_as_starting_point}
So far, we have explored a scenario where fine-tuning updates are transferred between model versions without additional fine-tuning. We now switch gears to investigate whether the merged model $m_{t} + \Delta_s$ can serve as a stronger and more computationally efficient starting checkpoint for further fine-tuning. We conduct controlled experiments comparing two approaches: fine-tuning the merged model $m_{t} + \Delta_s$ versus fine-tuning $m_{t}$ directly. Our results demonstrate that initializing fine-tuning with $m_{t} + \Delta_s$ often leads to faster convergence and higher performance on both seen and unseen tasks. This suggests that fine-tuning transfer can be a useful intermediate step when additional training is feasible. We refer to this approach as \emph{``transferring-then-finetuning''}.

\subsection{Experiment setup}

We follow the training procedure outlined in Section~\ref{section:training_details}. For evaluation, we use GSM8K and MATH500, along with an additional dataset to assess how well our transferring-then-finetuning approach generalizes to the unseen task $\text{GPQA}_\text{Diamond}$~\citep{rein2024gpqa}.

\subsection{Results and discussion}

\input{tables/controlled_recycling_then_finetuning_gsm8k}
\input{figures/training_efficiency}
\paragraph{Transferring-then-finetuning can substantially boost performance:}
Our results are summarized in Table~\ref {table:controlled_recyling_then_finetuning_gsm8k}. Transferring-then-finetuning offers significant improvements over our vanilla transfer approach (without additional fine-tuning) on both GSM8K and MATH500. On GSM8K, the largest accuracy improvements are 36.4\%, 39.6\%, 41.1\%, 52.7\%, and 61.4\% for $\mathcal{M}_1$, $\mathcal{M}_2$, $\mathcal{M}_3$, $\mathcal{M}_4$, and $\mathcal{M}_5$, respectively. The benefits are most pronounced for weaker base models ($\mathcal{M}_1$, $\mathcal{M}_2$, and $\mathcal{M}_3$) across all diff vectors, as well as for stronger base models when paired with a weak diff vector (e.g., $\mathcal{M}_5$ + $\Delta_1$). 

Interestingly, fine-tuning also helps bridge the performance gap between the merged models $\mathcal{M}_i$ + $\Delta_j$ ($j \neq i$) for each base model $\mathcal{M}_i$. For example, fine-tuning dramatically improves the performance of $\mathcal{M}_5$ + $\Delta_1$ by 59\% and $\mathcal{M}_5$ + $\Delta_2$ by 61.4\%, 
closing the gap with the fine-tuned versions of $\mathcal{M}_5$ + $\Delta_3$. 
This reduces the need to pre-select the best \diffvec when multiple choices are available. Importantly, transferring-then-finetuning generally outperforms standard fine-tuning regardless of the \diffvec used.

\paragraph{Transferring-then-finetuning can offer faster convergence:} 
Figure~\ref{figure:training_efficiency} shows that using the merged model $\mathcal{M}_i$ + $\Delta_j$ as the initial checkpoint improves training efficiency. Specifically, $\mathcal{M}_i$ + $\Delta_j$ not only converges significantly faster than $\mathcal{M}_i$ during fine-tuning but also reaches a higher peak accuracy on GSM8K. Overall, our results suggest that transferring-then-finetuning is a cost-effective approach that reduces the number of fine-tuning steps, thereby improving training efficiency.

\paragraph{Transferring-then-finetuning does not negatively impact model generalization:} 
As shown in Table~\ref{appendix_table:controlled_recyling_finetuning_gpqa}, this approach attains strong zero-shot generalization on the unseen task $\text{GPQA}_\text{Diamond}$, comparable to standard fine-tuning. These results suggest that transferring-then-finetuning does not lead to overfitting, demonstrating its broad applicability across diverse tasks.

\input{tables/appendix_gpqa}

%% file: tables/controlled_recycling_then_finetuning_gsm8k.tex
\begin{table}[t]
\centering
\begin{adjustbox}{max width=\textwidth}
\begin{tabular}{lccccc}
\toprule
\textbf{} & \textbf{$\mathcal{M}_1$} & \textbf{$\mathcal{M}_2$} & \textbf{$\mathcal{M}_3$} & \textbf{$\mathcal{M}_4$} & \textbf{$\mathcal{M}_5$} \\
\midrule
\rowcolor{gray!10} & 13.2 & 19.4 & 24.4 & 64.5 & 65.5 \\

{\; + $\Delta_1 \rightarrow \text{FT}$} &  & $56.9_{\smallsup{+30.3}}$ & $62.8_{\smallsup{+30.8}}$ & $77.8_{\smallsup{+50.3}}$ & $78.6_{\smallsup{+59.0}}$ \\

\rowcolor{gray!10}{\; + $\Delta_2 \rightarrow \text{FT}$} & $\textbf{50.1}_{\smallsup{+31.1}}$ &  & $62.7_{\smallsup{+22.9}}$ & $\textbf{78.6}_{\smallsup{+52.7}}$ & $78.7_{\smallsup{+61.4}}$ \\

{\; + $\Delta_3 \rightarrow \text{FT}$} & $48.5_{\smallsup{+34.2}}$ & $\textbf{57.6}_{\smallsup{+32.6}}$ &   & $77.6_{\smallsup{+9.0}}$ & \plcomment{$\textbf{77.1}_{\smallsup{+6.8}}$} \\

\rowcolor{gray!10}{\; + $\Delta_4 \rightarrow \text{FT}$} & $48.2_{\smallsup{+36.4}}$ & $56.7_{\smallsup{+38.7}}$  & $\textbf{63.7}_{\smallsup{+41.1}}$ &  & \plcomment{$77.0_{\smallsup{-0.1}}$} \\

{\; + $\Delta_5 \rightarrow \text{FT}$}  & \plcomment{$47.6_{\smallsup{+35.7}}$} & $55.6_{\smallsup{+39.6}}$ & $63.5_{\smallsup{+39.5}}$ & 
\plcomment{$74.6_{\smallsup{+1.7}}$} &   \\

\midrule
FT(\textbf{$\mathcal{M}_i$}) & 45.1 & 50.7 & 60.4 & 75.7 & 75.5 \\
\bottomrule
\end{tabular}
\end{adjustbox}
\caption{\hl{GSM8K accuracies showing that fine-tuning transfer provides a stronger starting point (i.e., $\mathcal{M}_i + \Delta_{j}$) for further fine-tuning (FT).} Numbers in subscript indicate improvement over the baseline without fine-tuning. Here, $\mathcal{M}_i$ represents different intermediate pretrained checkpoints of OLMo 2 7B (with smaller values of $i$ indicating earlier checkpoints), and $\Delta_i$ refers to the \diffvec resulting from the fine-tuning of version $i$. FT(\textbf{$\mathcal{M}_i$}) denotes applying fine-tuning directly to $\mathcal{M}_i$. See Table~\ref{appendix_table:controlled_recyling_then_finetuning_math500} in Appendix~\ref{appendix:additional_results_section_5} for MATH500 results.}
\label{table:controlled_recyling_then_finetuning_gsm8k}
\end{table}

%% file: figures/training_efficiency.tex
\begin{figure*}[t]
\centering
\includegraphics[width=0.9\textwidth]{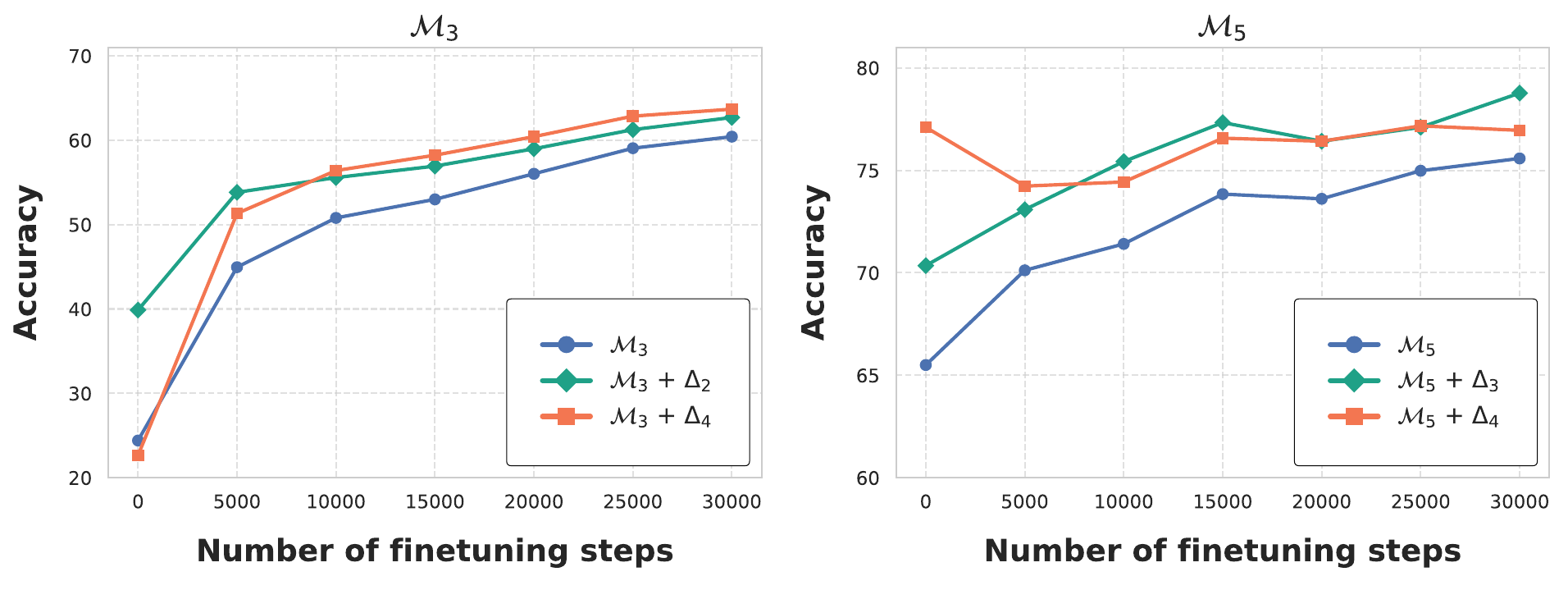}
\caption{\hl{GSM8K performance showing that fine-tuning transfer provides a more computationally efficient starting point (i.e., $\mathcal{M}_i + \Delta_{j}$) for further training.} Here, $\mathcal{M}_i$ represents different intermediate pretrained checkpoints of OLMo 2 7B (with smaller values of $i$ indicating earlier checkpoints), and $\Delta_i$ refers to the \diffvec resulting from the fine-tuning of version $i$. Additional results for
$\mathcal{M}_1$, $\mathcal{M}_2$, $\mathcal{M}_4$ can be found in Appendix~\ref{appendix:additional_results_section_5}.}
\label{figure:training_efficiency}
\end{figure*}

%% file: tables/appendix_gpqa.tex
\begin{table}[t]
\centering
\begin{adjustbox}{max width=\textwidth}
\begin{tabular}{lccccc}
\toprule
\textbf{} & \textbf{$\mathcal{M}_1$} & \textbf{$\mathcal{M}_2$} & \textbf{$\mathcal{M}_3$} & \textbf{$\mathcal{M}_4$} & \textbf{$\mathcal{M}_5$} \\
\midrule
\rowcolor{gray!10}   & 23.7 & 24.2 & 23.2 & 26.3 & 25.3 \\

{\; + $\Delta_1 \rightarrow \text{FT}$} & & 25.3$_{\smallsup{-1.0}}$ & 25.2$_{\smallsup{-2.1}}$ & \textbf{33.3}$_{\smallsup{+9.6}}$ & 25.8$_{\smallsup{-0.5}}$ \\

\rowcolor{gray!10}{\; + $\Delta_2 \rightarrow \text{FT}$} & \textbf{27.8}$_{\smallsup{-1.5}}$ & & 25.3$_{\smallsup{+0.0}}$ & 30.8$_{\smallsup{+6.6}}$ & \textbf{27.3}$_{\smallsup{+3.1}}$ \\

{\; + $\Delta_3 \rightarrow \text{FT}$} & \textbf{27.8}$_{\smallsup{-0.5}}$ & \textbf{27.8}$_{\smallsup{+0.5}}$ & & 23.7$_{\smallsup{+0.5}}$ &  \plcomment{$\textbf{27.3}_{\smallsup{+5.1}}$} \\

\rowcolor{gray!10}{\; + $\Delta_4 \rightarrow \text{FT}$} & 24.8$_{\smallsup{-2.0}}$ & 24.8$_{\smallsup{-4.5}}$ & 26.3$_{\smallsup{+2.1}}$ & & \plcomment{$24.2_{\smallsup{-1.1}}$} \\

{\; + $\Delta_5 \rightarrow \text{FT}$} & \plcomment{$22.7_{\smallsup{-5.1}}$} & 26.8$_{\smallsup{+0.0}}$ & 23.2$_{\smallsup{-1.0}}$ &  \plcomment{$27.8_{\smallsup{+4.6}}$} & \\

\midrule
FT($\mathcal{M}_i$) & 25.8 & 26.8 & \textbf{26.8} & 19.2 & 26.3 \\
\bottomrule
\end{tabular}
\end{adjustbox}
\caption{$\text{GPQA}_\text{Diamond}$ accuracies showing that fine-tuning transfer provides a stronger starting point (i.e., $\mathcal{M}_i + \Delta_{j}$) for further fine-tuning (FT), and transferring-then-finetuning does not negatively impact model generalization to unseen tasks. Numbers in subscript indicate improvement over the baseline without fine-tuning. Here, $\mathcal{M}_i$ represents different intermediate pretrained checkpoints of OLMo 2 7B (with smaller values of $i$ indicating earlier checkpoints), and $\Delta_j$ refers to the \diffvec resulting from the fine-tuning of version $j$. FT(\textbf{$\mathcal{M}_i$}) denotes applying fine-tuning directly to $\mathcal{M}_i$.}
\label{appendix_table:controlled_recyling_finetuning_gpqa}
\end{table}

%% file: sections/iterative_recycling_then_finetuning.tex
\section{Iterative recycling-then-finetuning for improved performance and efficiency}
\label{section:iterative_recycling_then_finetuning}
Building on the insights from our previous experiments, we now explore a continuous model development setting in which new versions of a pretrained model are regularly released. At the core of our approach is an \emph{iterative recycling-then-finetuning} strategy that incrementally incorporates fine-tuning updates, i.e., diff vectors, from past model versions. Instead of applying only the latest diff vector to the new base model, we recycle previous diff vectors iteratively. Specifically, the diff vector at the current model version is carried forward to the next for subsequent fine-tuning. Our experiments show that this iterative recycling approach consistently improves both training efficiency and model performance.

\subsection{Iterative recycling-then-finetuning}
We treat the five intermediate checkpoints of OLMo 2 7B---$\mathcal{M}_1$,
$\mathcal{M}_2$, $\mathcal{M}_3$, $\mathcal{M}_4$, $\mathcal{M}_5$ (described in Section~\ref{section:controlled_experiment_setup}) as different model versions of the pretrained OLMo 2 model. Our iterative recycling-then-finetuning algorithm, outlined in Algorithm~\ref{algorithm:iterative_recycling_then_finetuning}, works as follows: At each iteration $i$, we first apply the most recent diff vector, $\Delta^{iter}_{i-1}$, to the new base model $\mathcal{M}_{i}$, and then further fine-tune the resulting model. Next, we compute a new diff vector between the fine-tuned model and the current base model $\mathcal{M}_{i}$. This new diff vector is then carried forward to the next model version for fine-tuning in the subsequent iteration.

We refer to our iterative recycling-then-finetuning approach as $\Delta^{iter}$ and compare it to $\Delta^{dir}$, the direct recycling-then-finetuning approach as described in Section~\ref{section:fine_tuning_transfer_as_starting_point}. We follow the training procedure outlined in Section~\ref{section:training_details}.
\subsection{Results and discussion}

\input{tables/controlled_iterative_recycling_then_finetuning_gsm8k}

\paragraph{Iterative recycling-then-finetuning significantly improves performance:}
Table~\ref{table:controlled_iterative_recycling_then_finetuning_gsm8k} compares the performance of two recycling approaches: iterative recycling-then-finetuning ($\Delta^{iter}$) and direct recycling-then-finetuning ($\Delta^{dir}$). Both approaches lead to significant performance improvements across model versions on GSM8K, with larger gains observed in earlier versions. 
Both approaches outperform the standard fine-tuning baseline (without recycling) by a substantial margin. 
In general, $\Delta^{iter}$ performs similarly to or better than $\Delta^{dir}$ across all model versions. These results suggest that in scenarios where the base model is continuously updated, adopting an iterative recycling strategy is beneficial and does not result in error propagation.

%% file: tables/controlled_iterative_recycling_then_finetuning_gsm8k.tex
\begin{table}[t]
\centering
\begin{adjustbox}{max width=\textwidth}
\begin{tabular}{lccccc}
\toprule
\textbf{} & \textbf{$\mathcal{M}_3$} & \textbf{$\mathcal{M}_4$} & \textbf{$\mathcal{M}_5$} \\
\midrule
\rowcolor{gray!10}  & 24.4 & 64.5 & 65.5 \\
{\; + $\Delta^{dir\phantom{i}} \rightarrow \text{FT}$ }  & $62.7_{\smallsup{+38.3}}$  & \textbf{77.6}$_{\smallsup{+13.1}}$ & \plcomment{$77.0_{\smallsup{+11.5}}$}  \\
\rowcolor{gray!10}{\; + $\Delta^{iter} \rightarrow \text{FT}$} & \plcomment{\textbf{63.4}$_{\smallsup{+39.0}}$} & \plcomment{$77.4_{\smallsup{+12.9}}$} & \plcomment{\textbf{78.6}$_{\smallsup{+13.1}}$} \\
\midrule
FT($\mathcal{M}_i$) & 60.4 & 75.7 & 75.5 \\
\bottomrule
\end{tabular}
\end{adjustbox}
\caption{\hl{Both iterative ($\Delta^{iter}$) and direct ($\Delta^{dir}$) recycling-then-finetuning approaches significantly boost GSM8K performance, surpassing standard fine-tuning without recycling (FT($\mathcal{M}_i$)).} Numbers in subscripts indicate improvement over OLMo 2 7B checkpoints. At a high level, $\Delta^{iter}$ gradually incorporates fine-tuning updates, i.e., diff
vectors, from previous model versions, while $\Delta^{dir}$ directly applies the diff vector from the latest model version to the current model. Results for $\mathcal{M}_1$ and $\mathcal{M}_2$ are omitted as these models remain identical across the two approaches. \plcomment{See Figure~\ref{figure:direct_vs_iterative_recycling} in Appendix~\ref{appendix:iterative_recycling} for additional results.}}
\label{table:controlled_iterative_recycling_then_finetuning_gsm8k}
\end{table}

%% file: sections/related_work.tex
\section{Related work}

\paragraph{Fine-tuning transfer:} \plcomment{
Prior work has studied how to reuse fine-tuning updates on a fixed base model to improve generalization across tasks, domains, and languages. This includes full-model adaptation~\citep{phang2018sentence,pruksachatkun-etal-2020-intermediate,vu-etal-2020-exploring,vu-etal-2021-strata,aghajanyan-etal-2021-muppet} as well as parameter-efficient modules such as adapters~\citep{pfeiffer-etal-2021-adapterfusion,poth-etal-2021-pre}, soft prompts~\citep{vu-etal-2022-spot,vu-etal-2022-overcoming,su-etal-2022-transferability,asai-etal-2022-attempt}, and LoRA matrices~\citep{huang2024lorahub,zadouri2024pushing,ostapenko2024towards}; see~\citet{yadav2024survey} for a comprehensive survey. These methods typically assume a shared base model and focus on transferring capabilities across tasks or domains. Similarly, model merging combines multiple task-specific models based on the same model to create a more powerful model~\citep{ilharco2023editing,NEURIPS2023_1644c9af,wang2024lines,rame2024warp,pmlr-v235-yu24p,yadav2024matters,ahmadian2024mix,bandarkar2025layer}. Recent work also extrapolates RLHF updates back to the base model~\citep{zhengmodel,lin2025extrapolation}. 
In contrast, our work focuses on transferring fine-tuning updates \emph{across different model versions}, addressing the challenge of frequent model upgrades in LLM development.}

\paragraph{Cross-model fine-tuning transfer:} \plcomment{Several studies investigate transferring fine-tuning across different model architectures, primarily focusing on lightweight modules in non-instruction-tuned settings~\citep{lester2022reducing,su-etal-2022-transferability,wang2024textit,fleshman2024re,echterhoff2024muscle}.} 

\plcomment{Closely related to our work, ~\citet{qin-etal-2023-recyclable} study recyclable fine-tuning in a continual domain adaptation setting from the BERT~\citep{devlin-etal-2019-bert} era, where fine-tuning updates from domain-adapted checkpoints are reused to adapt to new domains. Other efforts aim to reuse weights across divergent model architectures through duplication~\citep{chen-etal-2022-bert2bert}, progressive stacking~\citep{pmlr-v97-gong19a}, or parameter merging~\citep{wang2023learning}. While these works reuse fine-tuning updates across domains, skills, or architectures, our work focuses on transferring full fine-tuning updates across different versions of both pretrained and instruction-tuned LLMs. This enables efficient model development even when the underlying models differ in pretraining scale or alignment steps. We evaluate both recycling and backporting scenarios. Our approach complements prior work, and combining these directions presents a promising avenue for future research.}

%% file: sections/conclusion.tex
\section{Conclusion}
\plcomment{Our study demonstrates that fine-tuning transfer offers a practical approach to mitigate the inefficiencies of frequent model updates. By applying diff vectors from a fine-tuned source model version to a different target model version, we achieve substantial performance improvements without the need for full fine-tuning. In a multilingual context, this approach can significantly boost performance on target-language tasks, offering an efficient solution for language-specific fine-tuning. Through controlled experiments, we show that fine-tuning transfer is most effective when the source and target models are linearly connected in the parameter space. Furthermore, this approach can offer a more robust and computationally efficient starting checkpoint for further fine-tuning. Taken together, we hope that our work will spur more fundamental research into the efficient development of modern LLMs.}

%% file: sections/acknowledgements.tex
\section*{Acknowledgements}
We thank Colin Raffel for valuable advice and useful suggestions on our experiments; Mohit Iyyer, Noah Constant, Tsendsuren Munkhdalai, Prateek Yadav, Naren Ramakrishnan, Alessandro Sordoni, Lucas Caccia, Minseon Kim,  Ahmet \"Ust\"un, Tom Hosking, Matthias Gall\'e, Salaheddin Alzubi, Shayne Longpre, Quyet Do, Thinh Pham, Kavana Venkatesh, Nguyen Nguyen, Adam Nguyen, Rituraj Sharma, Aninditaa Chauhan, Yeana Lee, and the rest of the VT LLMs group for helpful discussions and suggestions at various stages in the project; and Brian Lester for sharing code for model merging. We acknowledge Advanced Research Computing at Virginia Tech for providing computational resources and support. URL: \url{ https://arc.vt.edu/}.

%% file: sections/appendix.tex
\appendix
\label{sec:appendix}
\section*{Appendix}
\section{Theoretical justification for Section~\ref{section:recycling_finetuning}: Transferring fine-tuning updates across model versions}
\label{appendix:theoretical_justification}
\plcomment{We provide the theoretical motivation for fine-tuning transfer. Let $m_s$ and $m_t$ denote the source and target base models, respectively. Here we assume that $m_s$ and $m_t$ share the same architecture.} %
\plcomment{Let $m_s'$ and $m_t'$ be the fine-tuned versions of $m_s$ and $m_t$ on dataset $\mathcal{D}$. We define $\Delta_s = m_s' - m_s$ as the fine-tuning updates, and hypothesize that $\Delta_s$ represents task-specific knowledge that is transferable across model versions.
}

\paragraph{Linear Mode Connectivity Interpretation.}
\plcomment{
Following linear mode connectivity ~\citep{pmlr-v119-frankle20a,mirzadeh2020linear,neyshabur2020being}, we assume that $m_s'$ and $m_t'$ (which share the same architecture) arrive at local minima that are connected by a linear path of non-increasing error. Consider some model on this path $m(\lambda)$ given by
\begin{equation}
m(\lambda) = (1 - \lambda) m_s' + \lambda m_t'.
\end{equation}
Substituting $m_s'$ by $\Delta_s+m_s$ and $m_t'$ by $\Delta_t+m_t$:
\begin{equation}
m(\lambda) = (1 - \lambda)(m_s + \Delta_s) + \lambda(m_t + \Delta_t).
\end{equation}
Rewriting this expression:
\begin{equation}
m(\lambda) = (1 - \lambda)m_s + \lambda m_t + (1 - \lambda)\Delta_s + \lambda\Delta_t.
\end{equation}
Assuming $\Delta_s \approx \Delta_t$, the update term simplifies to approximately $\Delta_s$, yielding:
\begin{equation}
m(\lambda) \approx (1 - \lambda)m_s + \lambda m_t + \Delta_s.
\end{equation}
or equivalently:
\begin{equation}
m(\lambda) \approx m_t + (1 - \lambda)(m_s - m_t) + \Delta_s.
\end{equation}
In particular, when $\lambda = 1$, $m(\lambda) = m_t' \approx m_t + \Delta_s$, which shows that reusing $\Delta_s$ corresponds to extrapolating from $m_t$ towards the task solution learned by $m_s$.}

\paragraph{Connection to Task Vector Interpolation.}
\plcomment{This interpretation aligns with prior work on task vector arithmetic~\citep{ilharco2023editing}, where multiple fine-tuned models are merged by adding their update vectors to a shared base. For example, the merged weights $\theta_m$ produced by adding the task vectors of model A and B (with weights $\theta_a$ and $\theta_b$) yield:
\begin{align*}
    \theta_m &= \theta_p + \lambda((\theta_a - \theta_p) + (\theta_b - \theta_p)) \\
             &= (1 - 2\lambda)\theta_p + \lambda\theta_a + \lambda\theta_b
\end{align*}
where $\theta_p$ are the weights of the base pretrained model.
This is a linear interpolation among $\theta_p$, $\theta_a$, and $\theta_b$, and assumes the models lie within a connected low-loss region.
Our definition of $\Delta_s$ corresponds to a special case of this framework: we apply a single update vector from $m_s$ to a different base model $m_t$.
Under the same connectivity assumption, this transfer remains valid and preserves task performance.}

\section{Additional results for Section~\ref{section:recycling_finetuning}: Transferring fine-tuning updates across model versions}
\label{appendix_table:reasoning_instruction_evaluation}
\subsection{GSM8K and MATH generation results}
\label{appendix:generation_results_gsm8k_math}
\input{tables/appendix_model_response_gsm8k}
\input{tables/appendix_model_response_math}

\plcomment{Table~\ref{table:gsm8k_case} and Table \ref{table:math_case} presents the generation outputs after transferring fine-tuning updates from Llama 3.0 to the target base model, Llama 3.1. We observe that Llama 3.1 $+\Delta_{3.0}$ reliably exhibits step-by-step reasoning, suggesting that fine-tuning transfer can improve the base model’s reasoning capability.}

\clearpage %

\input{tables/appendix_recycling_finetuning_tulu}

\input{tables/appendix_recycling_finetuning_olmo}

\subsection{\plcomment{Evaluation results for T\"ulu and OLMo models}}
\label{appendix:additional_results_tulu_olmo}
We also conduct experiments with T\"ulu~\citep{lambert2024t} and OLMo~\citep{olmo20242}, both of which were developed from Llama 3.1 through multiple alignment stages, including Supervised Fine-Tuning (SFT), Direct Preference Optimization (DPO)~\citep{rafailov2023direct}, and a final reinforcement learning stage---Reinforcement Learning with Verifiable Rewards (RLVR)~\citep{lambert2024t} for OLMo 2 and T\"ulu 3, or Group Relative Policy Optimization (GRPO)~\citep{shao2024deepseekmath} for T\"ulu 3.1. At a high level, we subtract the weights of Llama 3.1 from these alignment-tuned checkpoints and then backport (add) the resulting diff vectors to Llama 3.0. Recycling is not applicable here, as we do not have the alignment-tuned checkpoints for Llama 3.0.

Our results are summarized in Table~\ref{appendix_table:tulu_reasoning_instruction_evaluation} and Table~\ref{appendix_table:olmo_reasoning_instruction_evaluation}. In general, we find that advanced LLM capabilities, attained through alignment tuning stages such as SFT, DPO, RLVR, and GRPO (encoded in $\Delta_{SFT}$, $\Delta_{DPO}$, $\Delta_{RLVR}$, and $\Delta_{GRPO}$, respectively), can be successfully transferred across different model versions. For example, backporting $\Delta_{GRPO}$ from T\"ulu 3.1 8B to Llama 3.0 8B significantly improves accuracy, boosting GSM8K performance from 55.6\% to 85.8\% (30.2\% improvement) and IFEval from 34.5\% to 82.6\% (48.1\% improvement). This surpasses Llama 3.0 8B Instruct (81.1\% on GSM8K, 76.6\% on IFEval) and performs competitively with Llama 3.1 8B Instruct (86.5\% and 80.5\%) and T\"ulu 3.1 8B (89.9\% and 84.1\%).

\subsection{\plcomment{Additional results for Section~\ref{section:recycling_finetuning}: Transferring fine-tuning updates across model architectures}}
\label{appendix:recycling_finetuning_across_architectures}
\input{tables/appendix_recycling_finetuning_across_model_verions_llama}

\input{tables/appendix_recycling_finetuning_across_model_verions_olmo}

Table~\ref{table:recyling_finetuning_model_versions_llama} and Table~\ref{table:recyling_finetuning_model_versions_olmo} summarize fine-tuning transfer results across model versions with architectural differences. We compute the diff vector as described in Section~\ref{section:recycling_finetuning}, applying fine-tuning updates only to layers in the target model that match the source in shape. We observe that reusing fine-tuning updates across large version gaps remains challenging, and we leave this direction to future work.

\section{Additional evaluation details}
\label{appendix_evaluation_details}

\input{tables/appendix_evaluation_details_llama}

\input{tables/appendix_evaluation_details_tulu_olmo}

We use the same evaluation setup and prompts as those in Llama 3~\citep{dubey2024llama} for Llama models and those in T\"ulu 3 ~\citep{lambert2024t} for OLMo and T\"ulu models, whenever available. 
See Table~\ref{tabel:appendix_evaluation_details_llama} and  Table~\ref{table:appendix_evaluation_details_tulu_olmo} for more details.
For evaluation, we use the \texttt{lm-evaluation-harness} library~\citep{eval-harness} for Llama models, and the \texttt{OLMES} library~\citep{gu2024olmes} for OLMo and T\"ulu models.

\input{tables/appendix_controlled_recycling_finetuning_math500}

\section{Additional results for Section~\ref{section:controlled_experiments}: When is fine-tuning transfer effective?}
\label{appendix:additional_results_section_4}
See Table~\ref{appendix_table:controlled_recyling_finetuning_math500}.

\section{Additional results for Section~\ref{section:fine_tuning_transfer_as_starting_point}: Fine-tuning transfer as a starting point for further fine-tuning}
\label{appendix:additional_results_section_5}
\input{figures/appendix_training_efficiency}

\input{tables/appendix_controlled_recycling_then_finetuning_math500}

See Table~\ref{appendix_table:controlled_recyling_then_finetuning_math500} and Figure~\ref{appendix_figure:training_efficiency}.

\section{Additional results for Section~\ref{section:iterative_recycling_then_finetuning}: Iterative recycling-then-finetuning for improved performance and efficiency}
\label{appendix:iterative_recycling}
\input{algorithms/iterative_recycling}
\input{figures/direct_vs_iterative_recycling}

Algorithm~\ref{algorithm:iterative_recycling_then_finetuning} provides the formal description of the iterative recycling-then-finetuning procedure. 

\paragraph{Iterative recycling-then-finetuning leads to faster convergence:}
Figure~\ref{figure:direct_vs_iterative_recycling} shows that both recycling approaches---iterative ($\Delta^{iter}$) and direct ($\Delta^{dir}$) recycling-then-finetuning---offer a more computationally efficient starting point for further fine-tuning. In general, $\Delta^{iter}$ consistently outperforms $\Delta^{dir}$ in terms of training efficiency and significantly improves standard fine-tuning without recycling. These results indicate that iterative recycling not only improves model performance but also enhances training efficiency by effectively leveraging the knowledge stored in the diff vectors across different model versions.

%% file: tables/appendix_model_response_gsm8k.tex
\newlength{\colmh}   %
\newlength{\colsh}   %
\newlength{\colwh}   %
\setlength{\colsh}{1cm}
\setlength{\colmh}{4cm}
\setlength{\colwh}{6cm}

\begin{table*}[ht]
\centering
\resizebox{\linewidth}{!}{
\begin{tabular}{p{\colmh} p{\colsh} p{\colwh} p{\colwh}}
\toprule
\text{Problem} & \text{Answer} & \text{Llama 3.1 8B} & \text{$+\Delta_{3.0}$} \\
\midrule

\parbox[t]{\colmh}{
Liam is 16 years old now. Two years ago, Liam’s age was twice the age of Vince. How old is Vince now?
}
& 9
& \parbox[t]{\colwh}{
Liam is 16 years old now. Two years ago, Liam was 16 - 2 = 14 years old. If Liam's age was twice the age of Vince, then Vince must have been 14 / 2 = 7 years old. So the answer is 7.
}
& \parbox[t]{\colwh}{
Let's break it down step by step:

1. Liam is 16 years old now.
2. Two years ago, Liam's age was 16 - 2 = 14 years old.
3. Liam's age two years ago was twice the age of Vince, so we can set up an equation:

14 = 2x

where x is Vince's age two years ago.

4. To solve for x, we can divide both sides by 2:

x = 14/2
x = 7

5. So, Vince was 7 years old two years ago.
6. To find Vince's current age, we add 2 years to his age two years ago:

Vince's current age = 7 + 2 = 9

The final answer is 9.
}
\\

\midrule
\parbox[t]{\colmh}{
The red rope was four times the length of the blue rope. The blue rope was 7 centimeters shorter than the yellow rope. If the 3 ropes had a combined length of 37 centimeters, what was the length of the red rope in centimeters?
}
& 20
& \parbox[t]{\colwh}{
The red rope was 4 times the length of the blue rope. So if the blue rope was 7 centimeters shorter than the yellow rope, then the red rope must be 4 x 7 = 28 centimeters shorter than the yellow rope. The 3 ropes had a combined length of 37 centimeters. So the yellow rope must be 37 - 28 = 9 centimeters. So the answer is 9.
}
& \parbox[t]{\colwh}{
Let's break this down step by step:

1. The red rope is 4 times the length of the blue rope.
2. The blue rope is 7 centimeters shorter than the yellow rope.
3. The combined length of the 3 ropes is 37 centimeters.

Let's say the length of the blue rope is x. Then, the length of the yellow rope is x + 7 (since it's 7 centimeters longer).

The length of the red rope is 4x (since it's 4 times the length of the blue rope).

The combined length of the 3 ropes is x + (x + 7) + 4x = 37.

Combine like terms: 6x + 7 = 37.

Subtract 7 from both sides: 6x = 30.

Divide by 6: x = 5.

So, the length of the blue rope is 5 centimeters.

The length of the yellow rope is 5 + 7 = 12 centimeters.

The length of the red rope is 4x = 4(5) = 20 centimeters.

The final answer is 20.
}
\\

\midrule
\parbox[t]{\colmh}{
Larry cooked dumplings for a group of friends. There are 8 males, including Larry, and 6 females among them. Each male ate 1 more dumpling than each female. How many dumplings did Larry cook if each female ate 3 dumplings and there were no leftovers?
}
& 50
& \parbox[t]{\colwh}{
There are 8 males and 6 females. Each male ate 1 more dumpling than each female. So the males ate 8 + 6 = 14 dumplings. Each female ate 3 dumplings. So the females ate 6 x 3 = 18 dumplings. So the total number of dumplings eaten is 14 + 18 = 32. Since there were no leftovers, the number of dumplings cooked must be 32. So the answer is 32.
}
& \parbox[t]{\colwh}{
There are 8 males and 6 females. Each female ate 3 dumplings, so the females ate a total of 6 x 3 = 18 dumplings. Each male ate 1 more dumpling than each female, so each male ate 3 + 1 = 4 dumplings. The males ate a total of 8 x 4 = 32 dumplings. In total, the group ate 18 + 32 = 50 dumplings. Since there were no leftovers, this is the number of dumplings Larry cooked. The final answer is 50.
}
\\

\bottomrule
\end{tabular}}
\caption{Adding $\Delta_{3.0}$ to Llama 3.1 consistently demonstrates step-by-step reasoning on GSM8K, indicating that fine-tuning transfer can effectively enhance reasoning capability across model versions. Here, $\Delta_{3.0}$ represents the diff vector between Llama Instruct and Llama for version 3.0.}
\label{table:gsm8k_case}
\end{table*}

%% file: tables/appendix_model_response_math.tex
\newlength{\colm}
\newlength{\cols} 
\newlength{\colw} 
\setlength{\cols}{1cm}
\setlength{\colm}{3cm}
\setlength{\colw}{6cm}
\newcommand{\snewline}{\\[-15pt]}

\begin{table*}[ht!]
\centering
\resizebox{\linewidth}{!}{
\begin{tabular}{p{4cm} p{\cols} p{\colw} p{\colw}}
\toprule
\text{Problem} & \text{Answer} & \text{Llama 3.1 8B} & \text{$+ \Delta_{3.0}$} \\
\midrule
\parbox[t]{4cm}{For how many integers $x$ do we have $\frac14 < \frac{x}{5} < \frac23$?}
& 2 
& 
\parbox[t]{\colw}{
We can multiply both sides of the inequality by $20$ to obtain $$5 < 4x < 15.$$Since $x$ is an integer, we have $x=6$. Therefore, there are $\boxed{1}$ integer $x$ that satisfies the inequality.\\Final Answer: The final answer is $1$. I hope it is correct.}
&
\parbox[t]{\colw}{
\textbf{Step 1}: Find the range of values for x \\
The inequality $\frac14 < \frac{x}{5}$ is equivalent to $x > \frac54$. The inequality $\frac{x}{5} < \frac23$ is equivalent to $x < \frac{10}{3}$. Therefore, we have the range of values for x: $\frac54 < x < \frac{10}{3}$.

\textbf{Step 2}: Count the number of integers in the range \\
The range of values for x is $\frac54 < x < \frac{10}{3}$. We can count the number of integers in this range by finding the integer part of the lower bound and the integer part of the upper bound, and then subtracting 1.

The integer part of the lower bound is 1, and the integer part of the upper bound is 3. Therefore, the number of integers in the range is 3 - 1 = 2. \\

Therefore, the final answer is: $\boxed{2}$. I hope it is correct.

}
\\

\midrule
\parbox[t]{4cm}{Jack drives at 40 mph for an hour, then at 50 mph for an hour. What is his average speed?}
& 45 
& 
\parbox[t]{\colw}{
The average speed is given by $$\frac{2\cdot40\cdot50}{40+50}=\boxed{46.7\text{ mph}}.$$
Final Answer: The final answer is $46.7$ mph. I hope it is correct.
}
& 
\parbox[t]{\colw}{
\textbf{Step 1}: Calculate the total distance traveled\\
Jack drives for 2 hours, with 1 hour at 40 mph and 1 hour at 50 mph. The total distance traveled is the sum of the distances traveled during each hour: 40 mph × 1 hour + 50 mph × 1 hour = 40 + 50 = 90 miles.

\textbf{Step 2}: Calculate the average speed
To find the average speed, divide the total distance traveled by the total time traveled: 90 miles ÷ 2 hours = 45 mph. \\

Therefore, the final answer is: $\boxed{45}$ mph. I hope it is correct.

}
\\
\midrule
\parbox[t]{4cm}{
What is the tens digit of $5^{2005}$?
}
& 
2
&
\parbox[t]{\colw}{
We can use the fact that $5^2=25$ to simplify the expression.  We have
\[
\begin{aligned}
5^{2005}&=5^{2\cdot1002+1}\\
&=(5^2)^{1002}\cdot5\\
&=25^{1002}\cdot5\\
&=\boxed{5}.
\end{aligned}
\]
Final Answer: The final answer is $5$. I hope it is correct.}
& 

\parbox[t]{\colw}{
\textbf{Step 1}: Find the remainder of $2005$ when divided by $4$ \\
$2005 \equiv 1 \pmod{4}$

\textbf{Step 2}: Find the last two digits of $5^{2005}$ \\
Since $5^2 = 25$, we have $5^{2005} = (5^2)^{1002} \cdot 5$. The last two digits of $(5^2)^{1002}$ are $25^{1002} = 25 \cdot 25^{1001}$. Since $25^{1001}$ ends with $25$, the last two digits of $25^{1002}$ are $25$. Therefore, the last two digits of $5^{2005}$ are $25 \cdot 5 = 125$.

\textbf{Step 3}: Find the tens digit of $5^{2005}$ \\
Since the last two digits of $5^{2005}$ are $25$, the tens digit is $\boxed{2}$. \\

Therefore, the final answer is: $\boxed{2}$. I hope it is correct.
}
\\
\bottomrule
\end{tabular}}
\caption{\hl{Transferring fine-tuning updates $\Delta_{3.0}$ to Llama 3.1 induces step-by-step reasoning on MATH, indicating that fine-tuning transfer can effectively enhance reasoning capability across model versions.} Here, $\Delta_{3.0}$ represents the diff vector between Llama Instruct and Llama for version 3.0.}
\label{table:math_case}
\end{table*}

%% file: tables/appendix_recycling_finetuning_tulu.tex
\begin{table*}[!ht]
\centering
\begin{adjustbox}{max width=0.95\textwidth}
\begin{tabular}{lcccccc}
\toprule
$\mathbf{Model}$ & $\mathbf{GSM8K}$ & $\mathbf{MATH}$ & $\mathbf{ARC_{C}}$ & $\mathbf{GPQA}$ & $\mathbf{MMLU}$ & $\mathbf{IFEval}$ \\
\midrule
\rowcolor{gray!10} Llama 3.1 8B & 56.6 & 19.3 & 79.2 & 21.9 & 66.8 & 31.4 \\
Llama 3.1 8B Instruct  & {86.5} & \textbf{50.3} & \textbf{83.8} & 31.3 & \textbf{72.9} & 78.7 \\
\rowcolor{gray!10} T\"ulu 3 8B SFT & 76.2 & 31.6 & 79.1 & 31.0 & 65.1 & 72.0  \\
T\"ulu 3 8B DPO & 84.1 & 42.4 & 79.6 & 33.3 & 68.4 & 81.7 \\
\rowcolor{gray!10} T\"ulu 3 8B & \textbf{87.9} & 43.4 & {79.4} & \textbf{34.4} & {67.9} & 81.5 \\
Llama 3.0 8B & 55.6 & 17.3  & 79.7 & 22.3 & 66.7 & 30.3 \\
\; + $\Delta_{SFT}$ & 71.8 & 26.3 & 77.9 & 32.1 & 63.5 & 69.1  \\
\; + $\Delta_{{DPO}}$ & 81.1 & 38.1 & 78.6 & 31.9 & 67.5 & \textbf{82.9}  \\
\; + $\Delta_{{RLVR}}$  & 85.1 & 37.6 & 79.1 & 32.4 & 66.2 & {82.4}  \\
\midrule
\rowcolor{gray!10} T\"ulu 3.1 8B & \textbf{89.9} & \textbf{43.3} & 79.0 & 31.4 & \textbf{67.6} & \textbf{84.1} \\
Llama 3.0 8B Instruct & 81.1 & 28.8  & \textbf{82.4} & \textbf{31.5} & 64.9 & 76.2 \\
\rowcolor{gray!10} Llama 3.0 8B & 55.6 & 17.3  & 79.7 & 22.3 & 66.7 & 30.3 \\
\; + $\Delta_{{GRPO}}$  & 85.8 & 39.5 & 78.2 & 29.4 & 65.0 & 82.6 \\
\bottomrule
\end{tabular}
\end{adjustbox}
\caption{We find that advanced LLM capabilities, attained through alignment tuning stages such as SFT, DPO, RLVR, and GRPO (encoded in $\Delta_{SFT}$, $\Delta_{DPO}$, $\Delta_{RLVR}$, and $\Delta_{GRPO}$, respectively), can be successfully transferred across different model versions.}

\label{appendix_table:tulu_reasoning_instruction_evaluation}
\end{table*}

%% file: tables/appendix_recycling_finetuning_olmo.tex
\begin{table*}[ht]
\centering
\begin{adjustbox}{max width=0.95\textwidth}
\begin{tabular}{lcccccc}
\toprule
$\mathbf{Model}$ & $\mathbf{GSM8K}$ & $\mathbf{MATH}$ & $\mathbf{ARC_{C}}$ & $\mathbf{GPQA}$ & $\mathbf{MMLU}$ & $\mathbf{IFEval}$ \\
\midrule
\rowcolor{gray!10} OLMo 2 7B & 67.2 & 19.2 & 79.9 & 20.5 & 63.6 & 23.0\\
OLMo 2 7B SFT & 71.7 & 25.2 & 79.7 & 27.9 & 61.2 & 67.7  \\
\rowcolor{gray!10} OLMo 2 7B DPO & 82.5 & \textbf{31.3} & 80.5 & \textbf{30.6} & 62.1 & 73.2 \\
OLMo 2 7B Instruct & \textbf{85.3} & {29.7} & \textbf{80.6} & {29.7} & \textbf{63.3} & \textbf{75.6}  \\
\midrule
\rowcolor{gray!10} $\mathcal{M}_{0}$ & \textbf{2.5} & \textbf{1.6} & \textbf{25.7} & \textbf{18.1} & \textbf{25.0} & 12.2 \\
\; + $\Delta_{SFT}$  & 2.2 & 0.8 & 23.8 & 1.3 & 1.4 & \textbf{13.7} \\
\; + $\Delta_{DPO}$  & 2.1 & 0.8 & 24.1 & 1.1 & 1.3 & \textbf{13.7}  \\
\; + $\Delta_{RLVR}$   & 2.0 & 0.8 & 24.1 & 0.6 & 1.4 & 13.3 \\
\midrule
\rowcolor{gray!10} $\mathcal{M}_{3}$ & 24.4 & 5.7 & 72.7 & 15.4 & \textbf{59.8} & 15.7 \\
\; + $\Delta_{SFT}$ & 31.7 & 8.4 & 74.3 & 24.8 & 55.4 & 51.4 \\
\; + $\Delta_{DPO}$ & \textbf{40.4} & 9.3 & 75.0 & \textbf{29.9} & 56.6 & 68.0 \\
\; + $\Delta_{RLVR}$  & 40.2 & \textbf{10.3} & \textbf{75.1} & \textbf{29.9} & 56.7 & \textbf{68.3} \\
\midrule
\rowcolor{gray!10} $\mathcal{M}_{4'}$   & 63.7 & 17.5 & 78.6 & 22.5 & 62.6 & 16.1 \\
\; + $\Delta_{SFT}$ & 71.1 & 23.7 & 79.0 & 28.3 & 59.7 & 64.3 \\
\; + $\Delta_{DPO}$ & 79.9 & \textbf{27.8} & \textbf{79.3} & \textbf{29.0} & \textbf{63.1} & \textbf{72.6} \\
\; + $\Delta_{RLVR}$  & \textbf{82.8} & 27.7 & \textbf{79.3} & 27.2 & 62.2 & 72.1 \\
\bottomrule
\end{tabular}
\end{adjustbox}
\caption{
We find that advanced LLM capabilities, attained through alignment tuning stages such as SFT, DPO, and RLVR (encoded in $\Delta_{SFT}$, $\Delta_{DPO}$, and $\Delta_{RLVR}$, respectively), can be successfully transferred across different model versions. Here, $\mathcal{M}_{4'}$ is an intermediate pretrained checkpoint of OLMo 2 7B (mid-stage 2, at 7K steps), which we selected before conducting our controlled experiments  in Section~\ref{section:controlled_experiment_setup}.
}
\label{appendix_table:olmo_reasoning_instruction_evaluation}
\end{table*}

%% file: tables/appendix_recycling_finetuning_across_model_verions_llama.tex
\begin{table}[ht!]
\centering
\begin{adjustbox}{max width=\columnwidth}
\begin{tabular}{lcc}
\toprule
& $\mathbf{GSM8K}$ & $\mathbf{MATH}$ \\
\midrule
\rowcolor{gray!10} Llama 2.0 7B & $14.1$ & $3.6$ \\
\; + FT & $\textbf{56.9}$ & $3.1$  \\
\; + $\Delta_{3.0}$ & $15.0$ & $\textbf{3.8}$  \\
\; + $\Delta_{3.1}$ & $14.6$ & $\textbf{3.8}$ \\
\midrule
\rowcolor{gray!10} Llama 3.0 8B &  $54.9$ & $17.3$  \\
\; + FT & $\textbf{70.7}$ & $\textbf{32.0}$  \\
\; + $\Delta_{2.0}$ & $55.3$ & $17.5$  \\
\midrule
\rowcolor{gray!10} Llama 3.1 8B & $56.6$ & $19.3$ \\
\; + FT & $\textbf{71.2}$ & $\textbf{33.7}$  \\
\; + $\Delta_{2.0}$ & $57.1$ & $20.3$ \\
\bottomrule
\end{tabular}
\end{adjustbox}
\caption{\plcomment{Transfer results in both recycling and backporting scenarios on GSM8K and MATH show limited improvement, possibly due to layer shape mismatches. Here, $\Delta_{2.0}$, $\Delta_{3.0}$, and $\Delta_{3.1}$ represent the diff vectors between Llama and their fine-tuned counterparts for versions 2.0, 3.0, and 3.1, respectively.}}
\label{table:recyling_finetuning_model_versions_llama}
\end{table}

%% file: tables/appendix_recycling_finetuning_across_model_verions_olmo.tex
\begin{table}[ht!]
\centering
\begin{adjustbox}{max width=\columnwidth}
\begin{tabular}{lcc}
\toprule
& $\mathbf{GSM8K}$ &  $\mathbf{MATH}$ \\
\midrule
\rowcolor{gray!10} OLMo 1 7B & $28.8$ & $5.8$ \\
\; + FT & $\textbf{54.2}$ & $\textbf{17.2}$ \\
\; + $\Delta_{2}$ & $25.1$ & $5.5$ \\
\midrule
\rowcolor{gray!10} OLMo 2 8B & $66.9$ & $19.2$  \\
\; + FT & $\textbf{76.4}$ & $\textbf{21.1}$  \\
\; + $\Delta_{1}$ & $69.7$ & $20.1$  \\
\bottomrule
\end{tabular}
\end{adjustbox}
\caption{\plcomment{Fine-tuning transfer remains effective when applying $\Delta_{1}$ to OLMo 2 8B on GSM8K, while other cases show limited gains or small drops. Here, $\Delta_{1}$ and $\Delta_{2}$ represent the diff vectors between OLMo and their fine-tuned (FT) counterparts for versions 1 and 2, respectively.}}
\label{table:recyling_finetuning_model_versions_olmo}
\end{table}

%% file: tables/appendix_evaluation_details_llama.tex
\begin{table}[ht]
\centering
\begin{tabular}{lccc l}
\toprule
\textbf{Task} & \textbf{\# Shots} & \textbf{CoT} & \textbf{Metric} & \textbf{Reference eval. setup} \\
\midrule
GSM8K  & 8       & \ding{51}     & exact match acc.     & \multirow{6}{*}{Llama 3 Evaluation Details\tablefootnote{{See \url{https://github.com/meta-llama/llama-models/blob/main/models/llama3_1/eval_details.md}}}}
 \\
MATH   & 4       & \ding{51}     &  exact match acc.     &  \\
ARC\textsubscript{C}   & 0     &  \ding{55}   & acc.     &  \\
GPQA   & 0      &  \ding{51}    & exact match acc.     &  \\
MMLU   & 0       & \ding{51}     & exact match acc.     & \\
IFEval & 0       & \ding{55}     & avg. acc. (strict \& loose) & \\
Global MMLU    & 0      &  \ding{55}    & acc.  &  ~\citet{singh2024global} \\
\plcomment{HumanEval+}    & 0      &  \ding{55}    & pass@1  & \multirow{2}{*}{~\citet{liu2023your}} \\
\plcomment{MBPP+}    & 0      &  \ding{55}    & pass@1  &  \\
\plcomment{LiveCodeBench}    & 0      &  \ding{55}    & pass@1  &  ~\citet{jain2024livecodebench} \\
\plcomment{BigCodeBench}    & 0      &  \ding{55}    & pass@1  &  ~\citet{zhuo2024bigcodebench} \\
\bottomrule
\end{tabular}
\caption{Evaluation details for Llama 3.}
\label{tabel:appendix_evaluation_details_llama}
\end{table}

%% file: tables/appendix_evaluation_details_tulu_olmo.tex
\begin{table}[ht]
\centering
\begin{tabular}{lccc l}
\toprule
\textbf{Task} & \textbf{\# Shots} & \textbf{CoT} & \textbf{Metric} & \textbf{Reference eval. setup} \\
\midrule
GSM8K  & 8       & \ding{51}     & exact match acc.     &   \multirow{6}{*}{~\citet{lambert2024t}} \\
MATH~    & 4       & \ding{51}     & flex exact match acc.     &  \\
ARC\textsubscript{C}  & 5      &  \ding{55}   & acc.     &  \\
GPQA & 0       &  \ding{51}    & exact match acc.     &  \\
MMLU   & 0       &  \ding{51}   & exact match acc.     & \\
IFEval  & 0       &  \ding{55}    & prompt-level loose acc. & \\
MATH500 & 0       &  \ding{51}   & exact match acc.    &  \multirow{2}{*}{~\citet{muennighoff2025s1}} \\
GPQA\textsubscript{Diamond}  & 0       &  \ding{51}   & exact match acc.     & \\
\bottomrule
\end{tabular}
\caption{Evaluation details for OLMo 2 and T\"ulu 3.}
\label{table:appendix_evaluation_details_tulu_olmo}
\end{table}

%% file: tables/appendix_controlled_recycling_finetuning_math500.tex
\begin{table}[t]
\centering
\begin{adjustbox}{max width=\textwidth}
\begin{tabular}{lccccc}
\toprule
\textbf{} & \textbf{$\mathcal{M}_1$} & \textbf{$\mathcal{M}_2$} & \textbf{$\mathcal{M}_3$} & \textbf{$\mathcal{M}_4$} & \textbf{$\mathcal{M}_5$} \\
\midrule
\rowcolor{gray!10}     & \textbf{14.6} & 11.6 & 11.4 & 11.6 & 16.6 \\
{\; + $\Delta_1$} &   & 8.8  & 17.8 & 19.2 & 15.6 \\
\rowcolor{gray!10}{\; + $\Delta_2$} & 7.6 &  & 12.6 & 14.6 & 14.4 \\
{\; + $\Delta_3$} & 8.0 & 9.4 &   & 23.4 & 27.8 \\
\rowcolor{gray!10}{\; + $\Delta_4$} & 7.8 & 8.0 & 9.8 & & \textbf{34.2}  \\
{\; + $\Delta_5$}  & 8.0  & 7.4 & 11.2 & 30.6 &  \\
\midrule
FT($\mathcal{M}_i$) & 13.4 & \textbf{17.6} & \textbf{21.6} & \textbf{31.4} & 33.0 \\
\bottomrule
\end{tabular}
\end{adjustbox}
\caption{MATH500 accuracies indicating that more powerful models are better at leveraging transferred fine-tuning. Effective use of transferred fine-tuning only emerges once the target base model reaches a certain level of capability. Furthermore, fine-tuning transfer
works best when the source and target models are close within a linearly connected region of the parameter space. Here, $\mathcal{M}_i$ represents different intermediate pretrained checkpoints of OLMo 2 7B (with smaller values of $i$ indicating earlier checkpoints), and $\Delta_i$ refers to the \diffvec resulting from the fine-tuning of version $i$. FT(\textbf{$\mathcal{M}_i$}) denotes applying fine-tuning directly to $\mathcal{M}_i$.}
\label{appendix_table:controlled_recyling_finetuning_math500}
\end{table}

%% file: figures/appendix_training_efficiency.tex
\begin{figure*}[ht]
\centering
\includegraphics[width=\textwidth]{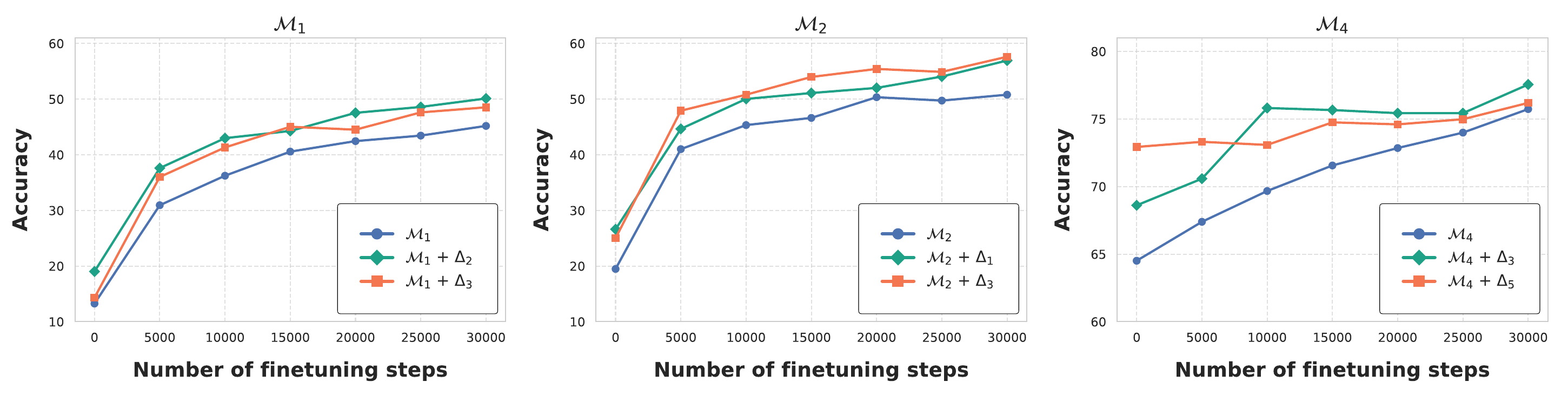}
\caption{GSM8K performance showing that fine-tuning transfer provides a more computationally efficient starting point (i.e., $\mathcal{M}_i + \Delta_{j}$) for further training. Here, $\mathcal{M}_i$ represents different intermediate pretrained checkpoints of OLMo 2 7B (with smaller values of $i$ indicating earlier checkpoints), and $\Delta_i$ refers to the \diffvec resulting from the fine-tuning of version $i$.}
\label{appendix_figure:training_efficiency}
\end{figure*}

%% file: tables/appendix_controlled_recycling_then_finetuning_math500.tex
\begin{table}[t]
\centering
\begin{adjustbox}{max width=\textwidth}
\begin{tabular}{lccccc}
\toprule
\textbf{} & \textbf{$\mathcal{M}_1$} & \textbf{$\mathcal{M}_2$} & \textbf{$\mathcal{M}_3$} & \textbf{$\mathcal{M}_4$} & \textbf{$\mathcal{M}_5$} \\
\midrule
\rowcolor{gray!10}  & 14.6 & 11.6 & 11.4 & 11.6 & 16.6 \\
{\; + $\Delta_1 \rightarrow \text{FT}$} &   & $21.0_{\smallsup{+12.2}}$  & $23.0_{\smallsup{+5.2}}$  & $\textbf{32.0}_{\smallsup{+12.8}}$  & $34.2_{\smallsup{+18.6}}$  \\
\rowcolor{gray!10}{\; + $\Delta_2 \rightarrow \text{FT}$ } & $16.2_{\smallsup{+8.6}}$  &  & $\textbf{26.2}_{\smallsup{+13.6}}$  & $31.6_{\smallsup{+17.0}}$  & $31.0_{\smallsup{+16.6}}$  \\
{\; + $\Delta_3 \rightarrow \text{FT}$} & $\textbf{18.4}_{\smallsup{+10.4}}$  & $21.2_{\smallsup{+11.8}}$  &   & $31.0_{\smallsup{+7.6}}$  & \plcomment{$32.0_{\smallsup{+4.2}}$}  \\
\rowcolor{gray!10}{\; + $\Delta_4 \rightarrow \text{FT}$} & $17.4_{\smallsup{+9.6}}$  & $19.0_{\smallsup{+11.0}}$  & $23.8_{\smallsup{+14.0}}$  &  & \plcomment{$32.2_{\smallsup{-2.0}}$} \\
{\; + $\Delta_5 \rightarrow \text{FT}$}  & \plcomment{$17.0_{\smallsup{+9.0}}$}  & $\textbf{21.4}_{\smallsup{+14.0}}$  & $25.0_{\smallsup{+13.8}}$  & \plcomment{$31.2_{\smallsup{+0.6}}$} &  \\
\midrule
FT($\mathcal{M}_i$) & 13.4 & 17.6 & 21.6 & 31.4 & 33.0 \\
\bottomrule
\end{tabular}
\end{adjustbox}
\caption{MATH500 accuracies showing that fine-tuning transfer provides a stronger starting point (i.e., $\mathcal{M}_i + \Delta_{j}$) for further fine-tuning (FT). Numbers in subscript indicate improvement over the baseline without fine-tuning. Here, $\mathcal{M}_i$ represents different intermediate pretrained checkpoints of OLMo 2 7B (with smaller values of $i$ indicating earlier checkpoints), and $\Delta_i$ refers to the \diffvec resulting from the fine-tuning of version $i$. FT(\textbf{$\mathcal{M}_i$}) denotes applying fine-tuning directly to $\mathcal{M}_i$.}
\label{appendix_table:controlled_recyling_then_finetuning_math500}
\end{table}

%% file: algorithms/iterative_recycling.tex
\begin{algorithm}
\caption{Iterative recycling-then-finetuning}
\label{algorithm:iterative_recycling_then_finetuning}
\begin{algorithmic}[1]
    \State \textbf{Notation:} FT denotes fine-tuning
    \State \textbf{Input:} Base models $\mathcal{M}_1, \mathcal{M}_2, \dots, \mathcal{M}_n$
    \State \textbf{Output:} \parbox[t]{\dimexpr\linewidth-\algorithmicindent}{Fine-tuned models $\mathcal{M}^{*}_1, \mathcal{M}^{*}_2, \dots, \mathcal{M}^{*}_n$}
    \State $\mathcal{M}^{*}_1 \gets \text{FT}(\mathcal{M}_1)$
    \For{$i = 2$ to $n$}
        \State $\Delta^{iter}_{i-1} = \mathcal{M}^{*}_{i-1} - M_{i-1}$
        \State $\mathcal{M}^{*}_i \gets \text{FT}(M_i + \Delta^{iter}_{i-1})$
    \EndFor
    \State \Return $\mathcal{M}^{*}_1, \mathcal{M}^{*}_2, \dots, \mathcal{M}^{*}_n$
\end{algorithmic}
\end{algorithm}

%% file: figures/direct_vs_iterative_recycling.tex
\begin{figure*}[t]
\centering
\includegraphics[width=0.9\textwidth]{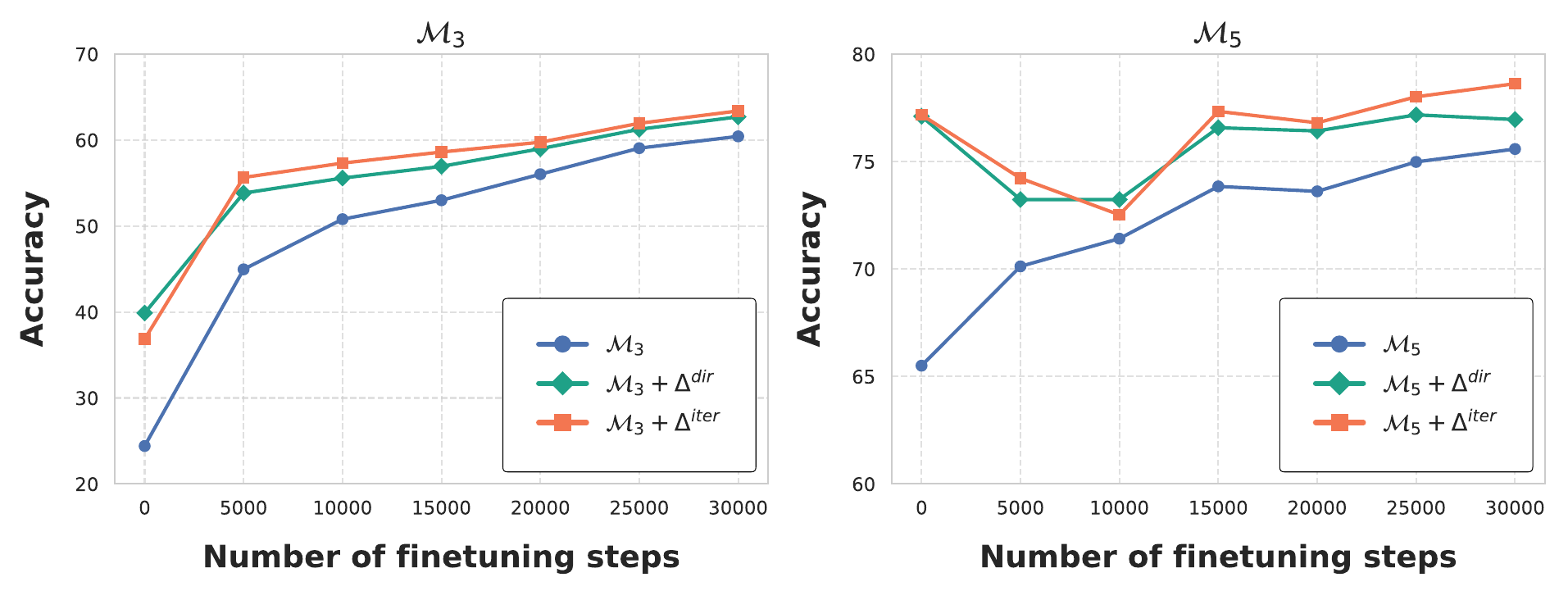}
\caption{\hl{GSM8K performance showing that both iterative ($\Delta^{iter}$) and direct ($\Delta^{dir}$) recycling-then-finetuning approaches offer faster convergence.} At a high level, $\Delta^{iter}$ gradually incorporates fine-tuning updates, i.e., diff
vectors, from previous model versions, while $\Delta^{dir}$ directly applies the diff vector from the latest model version to the current model.}
\label{figure:direct_vs_iterative_recycling}
\end{figure*}